\documentclass{article}
\pdfoutput=1
\usepackage{times}
\usepackage{graphicx} 
\usepackage{subfigure} 

\usepackage{natbib}

\usepackage{algorithm}
\usepackage{algorithmic}

\usepackage{mathtools}

\usepackage[accepted, nohyperref]{icml2016}
\usepackage{epstopdf}

\icmltitlerunning{Anomaly Detection in Video with Bayesian Nonparametrics}

\begin{document} 

\twocolumn[
\icmltitle{Anomaly Detection in Video\\ 
           with Bayesian Nonparametrics}

\icmlauthor{Olga Isupova}{o.isupova@sheffield.ac.uk}
\icmlauthor{Danil Kuzin}{dkuzin1@sheffield.ac.uk}
\icmlauthor{Lyudmila Mihaylova}{l.s.mihaylova@sheffield.ac.uk}
\icmladdress{The University of Sheffield,\\
            Western Bank, Sheffield, S10 2TN, UK}

\icmlkeywords{anomaly detection, Bayesian nonparametrics, topic modeling}

\vskip 0.3in
]

\begin{abstract} 
A novel dynamic Bayesian nonparametric topic model for anomaly detection in video is proposed in this paper. Batch and online Gibbs samplers are developed for inference. The paper introduces a new abnormality measure for decision making. The proposed method is evaluated on both synthetic and real data. The comparison with a non-dynamic model shows the superiority of the proposed dynamic one in terms of the classification performance for anomaly detection. 
\end{abstract} 

\section{Introduction}
\label{introduction}

Topic modeling~\cite{Hofmann99, Blei03LDA} is a promising approach for anomaly detection in video~\cite{Jeong14, Varadarajan2009, Mehran09}. This is an unsupervised method which means that there is no need to predict all kinds of abnormalities in advance and collect a labelled dataset for it. Topic modeling also provides additional information about typical motions and behaviours rather than just warns about abnormal events.

In the text mining application a topic model represents unlabelled documents as mixtures of topics where unknown topics are distributions over observed words. In conventional topic modeling documents are assumed to be independent. Although in some cases this assumption is not valid and different dynamic models are proposed in the literature~\cite{Blei2006Dynamic, Ahmed2010, Hospedales2011, Kuettel2010, PruteanuMalinici2010, Srebro2005, Zhang2010}.  

In the video processing application short video clips are often treated as documents, local motion patterns are represented by topics. All motions in the real life last for some time hence topic mixtures in the successive documents are expected to be similar if the clips are sufficiently short.

We propose a dynamic nonparametric topic model for anomaly detection. Successive documents are encouraged to have similar topic mixtures.  

Anomaly detection is an urgent task; the decision should be made as soon as possible. Batch and online Gibbs sampler is proposed in this paper. The online inference algorithm allows to estimate parameters for the current document with no need to rerun it on the previous ones.
An abnormality measure for decision making is also proposed in the paper.

The paper is organised as follows. Section~\ref{sec:video_rep} defines visual words and documents while section~\ref{sec:proposed_model} describes the proposed model. The inference and the whole framework are introduced in sections~\ref{sec:inference} and~\ref{sec:abnormality} respectively. Evaluation of the method using synthetic and real data is presented in section~\ref{sec:experiments} followed by the conclusions in section~\ref{sec:conclusions}.

\section{Visual Features}
\label{sec:video_rep}

\begin{figure}[ht]
\vskip 0.2in
\begin{center}
\centerline{\includegraphics[scale=0.2]{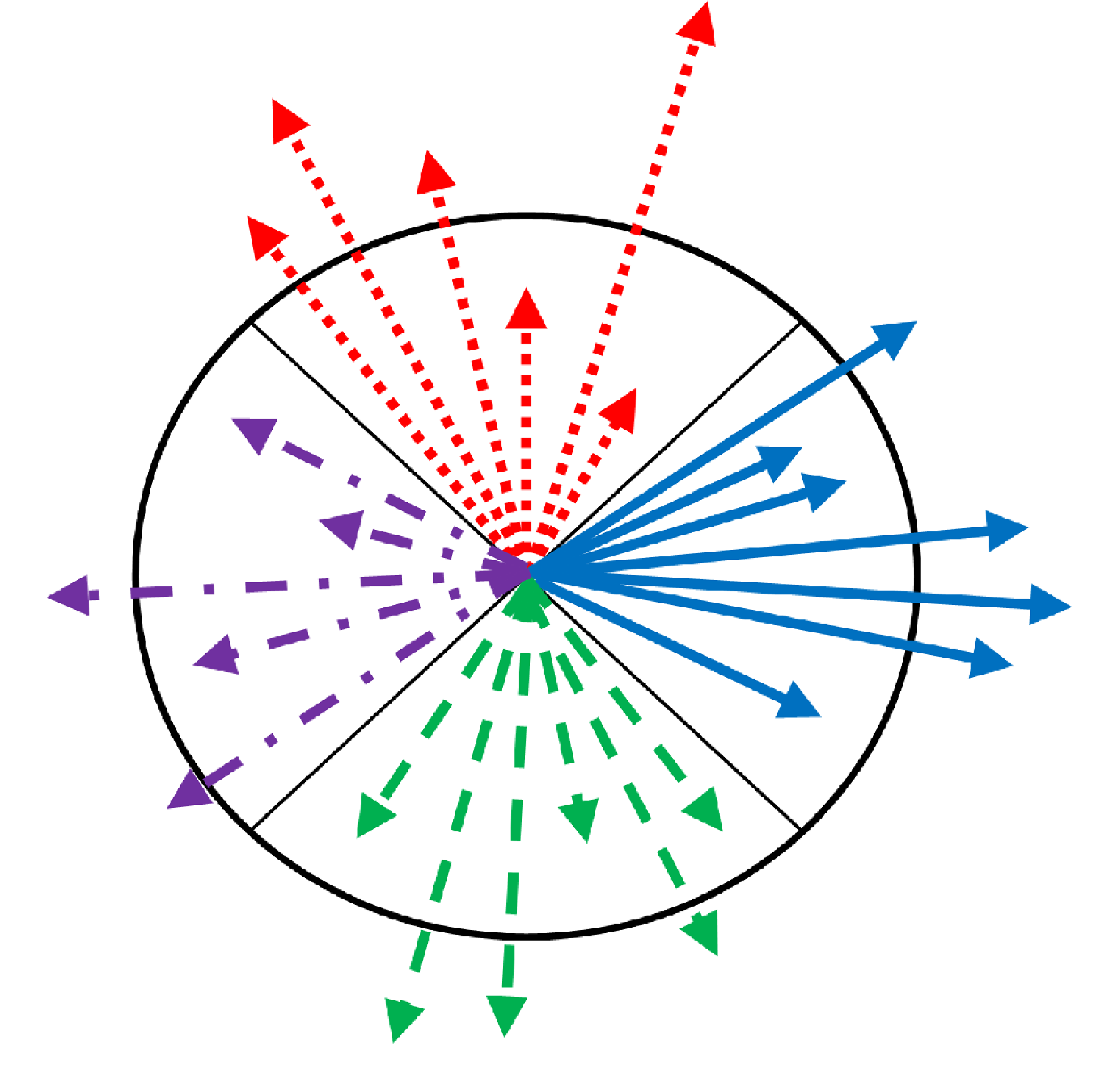}}
\caption{Optical flow direction quantisation. The four main directions are extracted --- up, right, down and left, highlighted by the colour on the figure.}
\label{fig:word_formation}
\end{center}
\vskip -0.2in
\end{figure}

The definitions of visual words and visual documents are essential for topic modeling application to video processing. A quantised direction (Figure~\ref{fig:word_formation}) of an average optical flow vector~\cite{Horn1981} over $N \times N$ pixels and its location form a visual word. Non-overlapping clips of the whole video sequence are treated as visual documents.  

\section{Proposed Model}
\label{sec:proposed_model}
Let $\mathbf{x}_{1:J} = \{\mathbf{x}_j\}_{j = 1 : J}$ denote a sequence of documents. A document $\mathbf{x}_j$ consists of $N_j$ words $x_{ji}$: $\mathbf{x}_j = \{x_{ji}\}_{i = 1:N_j}$. Documents are assumed to be mixtures of \textit{topics} $\{\boldsymbol\phi_k\}_{k = 1 : \infty}$, which are the latent distribution over words. The number of possible topics is expected to go to infinity for an infinite amount of data.

\subsection{Hierarchical Dirichlet Process Topic Model}
A hierarchical Dirichlet process (HDP)~\cite{Teh2012} can represent mixture models with  a potentially infinite number of mixture components.

A HDP can be represented in different ways, Chinese restaurant franchise (CRF) is reviewed here. Documents are considered as ``restaurants'' and words are considered as ``customers'' in this metaphor. The words in the documents form groups around ``tables'' and eat the same ``dish'', which corresponds to a topic, on one table. The set of the topics is shared among all the documents, that creates a ``franchise'' of the restaurants.   

Denote a table assignment of the token $i$ in the document $j$ by $t_{ji}$, a topic assignment of the table $t$ in the document $j$ by $k_{jt}$. The following counts are used: $n_{jt}$ for the number of words assigned to the table $t$ in the document $j$ and $m_{jk}$ for the number of tables serving the topic $k$ in the document $j$. Let dots in subscripts denote marginalisation over the corresponding dimension. 

The generative process is the following. Each token $i$ in the document $j$ chooses one of the occupied tables with a probability proportional to a number $n_{jt}$ of words already assigned to a table, or the token starts a new table with a probability proportional to a parameter $\alpha$:
\begin{equation}
\label{eq:hdp_t}
p(t_{ji} = t | t_{j1}, \dotsc, t_{j i-1}, \alpha) = 
\begin{cases}
\dfrac{n_{jt}}{i - 1 + \alpha}, \text{ if } t = 1 : m_{j\cdot}; \\[1.5mm]
\dfrac{\alpha}{i - 1 + \alpha}, \text{ if } t = t^{\text{new}}.
\end{cases}
\end{equation}

If a new table is started the topic should be assigned to it. It can be one of the used topics with a probability proportional to a number $m_{\cdot k}$ of tables having this topic among all the documents, or it can be a new topic with a probability proportional to a parameter $\gamma$:
\begin{equation}
\label{eq:hdp_k}
p(k_{j t^{\text{new}}} = k | k_{11}, \dotsc, k_{jt-1}, \gamma) =
\begin{cases} 
\dfrac{m_{\cdot k}}{m_{\cdot \cdot} + \gamma}, \text{ if } k = 1 : K; \\[1.5mm]
\dfrac{\gamma}{m_{\cdot \cdot} + \gamma}, \text{ if } k = k^{\text{new}},
\end{cases}
\end{equation} 
where $K$ is a number of topics used so far. In the case of a new topic it is sampled from the base measure $H$.

The word $x_{ji}$ for the topic $i$ in the document $j$ assigned to the table $t_{ji}$ is sampled from the topic $k_{j t_{ji}}$ served on this table:
\begin{equation}
\label{eq:hdp_x}
x_{j t} \sim \text{Mult}(\boldsymbol{\phi}_{k_{j t_{j i}}}).
\end{equation}

\subsection{Dynamic Hierarchical Dirichlet Process Topic Model}

Exchangeability of documents and words is an essential assumption in the HDP, which means that the joint probability of the data is independent of the order of the documents and words. Although in video processing this assumption is not reasonable. Motions last for some time and it is expected that the topic mixture of the current document is similar to the topic mixture in the previous one. However, the words inside documents are still exchangeable.

The dynamic extension of the HDP topic model is proposed in this paper to take into account this intuition. The probability of the topic $k$ being assigned to one of the tables in the document $j$ explicitly depends on the usage of this topic in the current and previous documents $m_{j k} + m_{j-1 k}$. The topic distribution of the current document is hence encouraged to be similar to the topic distribution of the previous one. 

The proposed model assumes the following generative process. A table assignment for a token remains unchanged~(\ref{eq:hdp_t}). A topic for a new table in the document $j$ is assigned to one of the used topics $k$ with a probability proportional to the sum of the number of tables serving this topic in the current and previous documents $m_{jk} + m_{j-1 k}$ and the weighted number of tables among all the documents that have this topic $\delta \, m_{\cdot k}$, where $\delta$ is a parameter of the model, or it is assigned to a new topic with a probability proportional to the parameter $\gamma$: 
\begin{align}
&p(k_{jt} = k | k_{11}, \dotsc, k_{jt-1}, \gamma) = \nonumber\\
\label{eq:dynHDP_k}
&\begin{cases}
\dfrac{m_{jk} + m_{j-1 k} + \delta m_{\cdot k}}{m_{j \cdot} + m_{j-1 \cdot} + \delta m_{\cdot \cdot} + \gamma}, \text{ if } k = 1:K;\\[1.5mm]
\dfrac{\gamma}{m_{j\cdot} + m_{j-1 \cdot} + \delta m_{\cdot \cdot} + \gamma}, \text{ it } k = k^{\text{new}}.
\end{cases}
\end{align}

The word $x_{j i}$ is sampled as in the HDP from the corresponding topic as defined in (\ref{eq:hdp_x}).

\section{Inference}
\label{sec:inference}
Conventional inference algorithms are batch algorithms, i.e. they process the whole dataset, which is computationally intractable for large or stream datasets. Online algorithms work sequentially, one data point at a time. We propose a combination of offline and online inference for our model.

We use Gibbs sampling~\cite{Geman1984}. The hidden variables $\mathbf{t} = \{t_{j i}\}_{j = 1 : J, i = 1 : N_j}$ and $\mathbf{k} = \{k_{j t}\}_{j = 1 : J, t = 1 : m_{j \cdot}}$ are sampled from their conditional distributions.

The batch Gibbs sampler is run for the training set of the documents. After this training stage the global estimates of the topics $\boldsymbol{\phi}_k$ and the counts $m_{\cdot k}$ for all $k$ are stored and used for the online inference of the testing documents. For each testing document the online Gibbs sampler is run to sample table assignments and topic assignments for this document only. The online Gibbs sampler updates the local counts $m_{jk}$. After the Gibbs sampler converges, the global variables $\boldsymbol{\phi}_k$ and $m_{\cdot k}$ are updated with the information obtained by the new document. 
\begin{table}[t]
\caption{AUC results.}
\label{tab:auc}
\begin{center}
\begin{small}
\begin{sc}
\begin{tabular}{cccc}
\hline
\abovespace\belowspace
Data set & Dynamic HDP & HDP & ``True'' model \\
\hline
\abovespace
Synthetic & 0.7118 & 0.4751 & 0.7280\\
\belowspace
QMUL & 0.7100 & 0.4644 & ---\\
\hline
\end{tabular}
\end{sc}
\end{small}
\end{center}
\vskip -0.1in
\end{table}
\section{Anomaly Detection}
\label{sec:abnormality}
Anomaly detection can be done within the probabilistic framework using topic modeling. In this framework the data point is assumed to be abnormal if it has a low value of likelihood, i.e. the learnt model cannot explain the current observation because something atypical is happening. We use the predictive likelihood estimated as a harmonic mean~\cite{Griffiths2004} and normalised by the length~$N_j$ of the document for anomaly detection. 

\section{Experiments}
\label{sec:experiments}
The proposed method\footnote{https://github.com/OlgaIsupova/dynamic-hdp} is applied for anomaly detection on synthetic and real data. We compare it with the method based on the HDP topic model  (for the batch Gibbs sampler of the HDP topic model the implementation by Chong Wang is used\footnote{https://github.com/Blei-Lab/hdp}). For the quantitative comparison the area (AUC) under the receiver operating characteristic (ROC) curve~\cite{Murphy2012} for abnormality classification accuracy is used. 

\subsection{Synthetic Data}
\begin{figure}[!t]
\vskip 0.2in
\begin{center}
\centerline{\includegraphics[width=0.95\columnwidth]{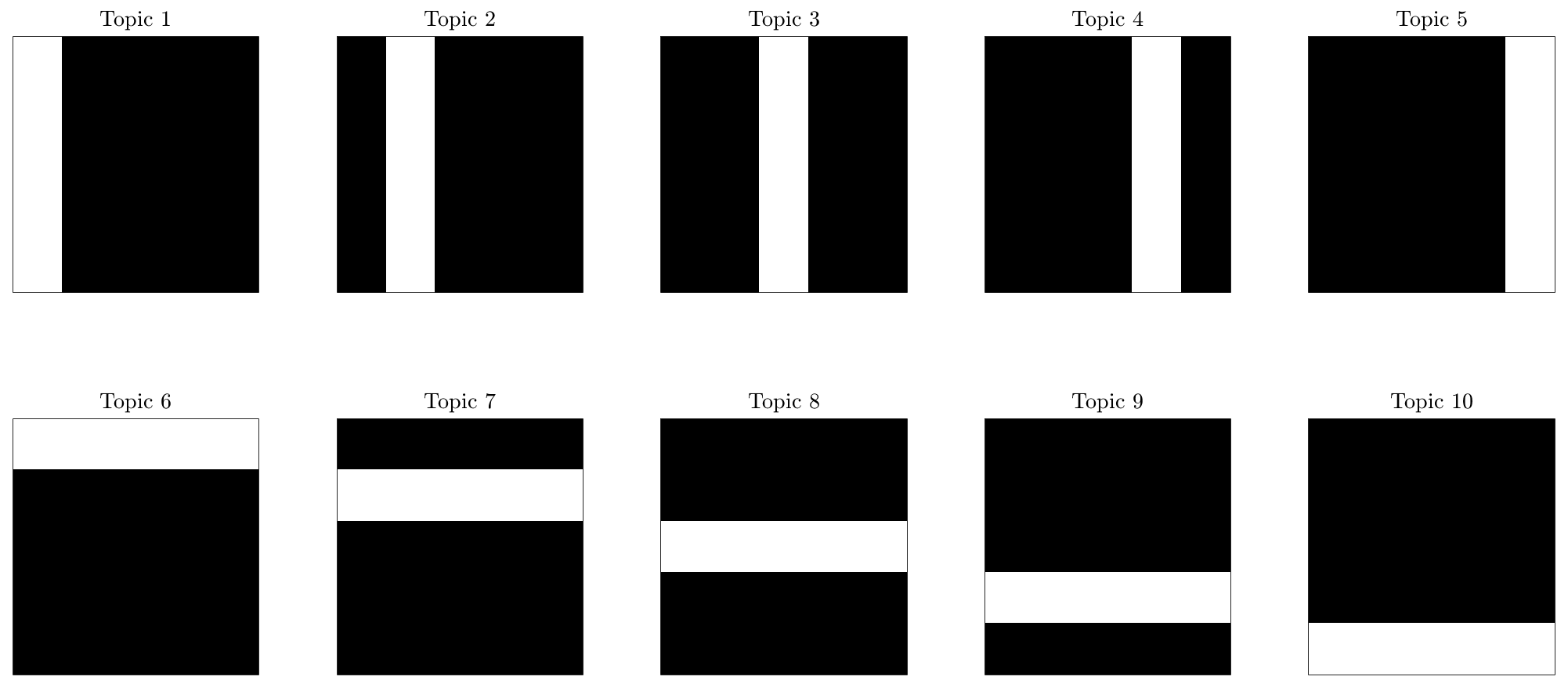}}
\caption{Graphical representation of the topics in the synthetic dataset. There are $25$ words, organised into a $5 \times 5$ matrix, where a word corresponds to a cell in this matrix. The topics are represented as the coloured matrices, where the colour of the cell indicates the probability of the corresponding word in a given topic, the lighter the colour the higher the probability value is.}
\label{fig:synthetic_topics}
\end{center}
\vskip -0.2in
\end{figure}

The ``bar'' data introduced in~\cite{Griffiths2004} is used. The vocabulary consists of $V = 25$ words, organised into a $5 \times 5$ matrix. The topics $\{\boldsymbol\phi_k\}_{k = 1}^{10}$ form vertical and horizontal bars in the matrix (Figure~\ref{fig:synthetic_topics}). 

\begin{figure}[!t]
\vskip 0.2in
\begin{center}
\centerline{\includegraphics[scale=0.4]{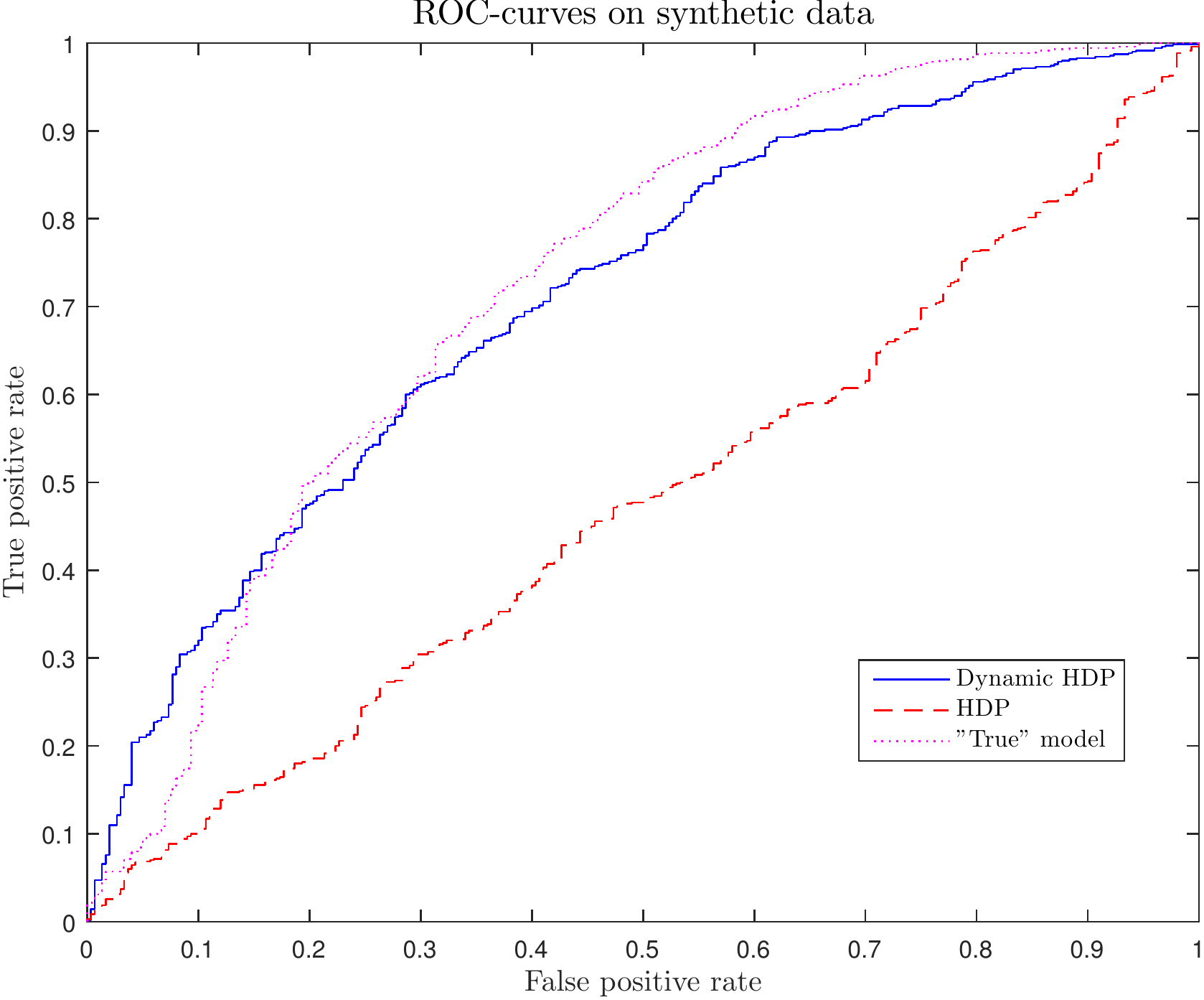}}
\caption{The ROC-curves for the synthetic data.}
\label{fig:synthetic_roc}
\end{center}
\vskip -0.2in
\end{figure} 

\begin{figure*}[!t]
\vskip 0.2in
\begin{center}
\subfigure[]{\includegraphics[width = 0.47\columnwidth]{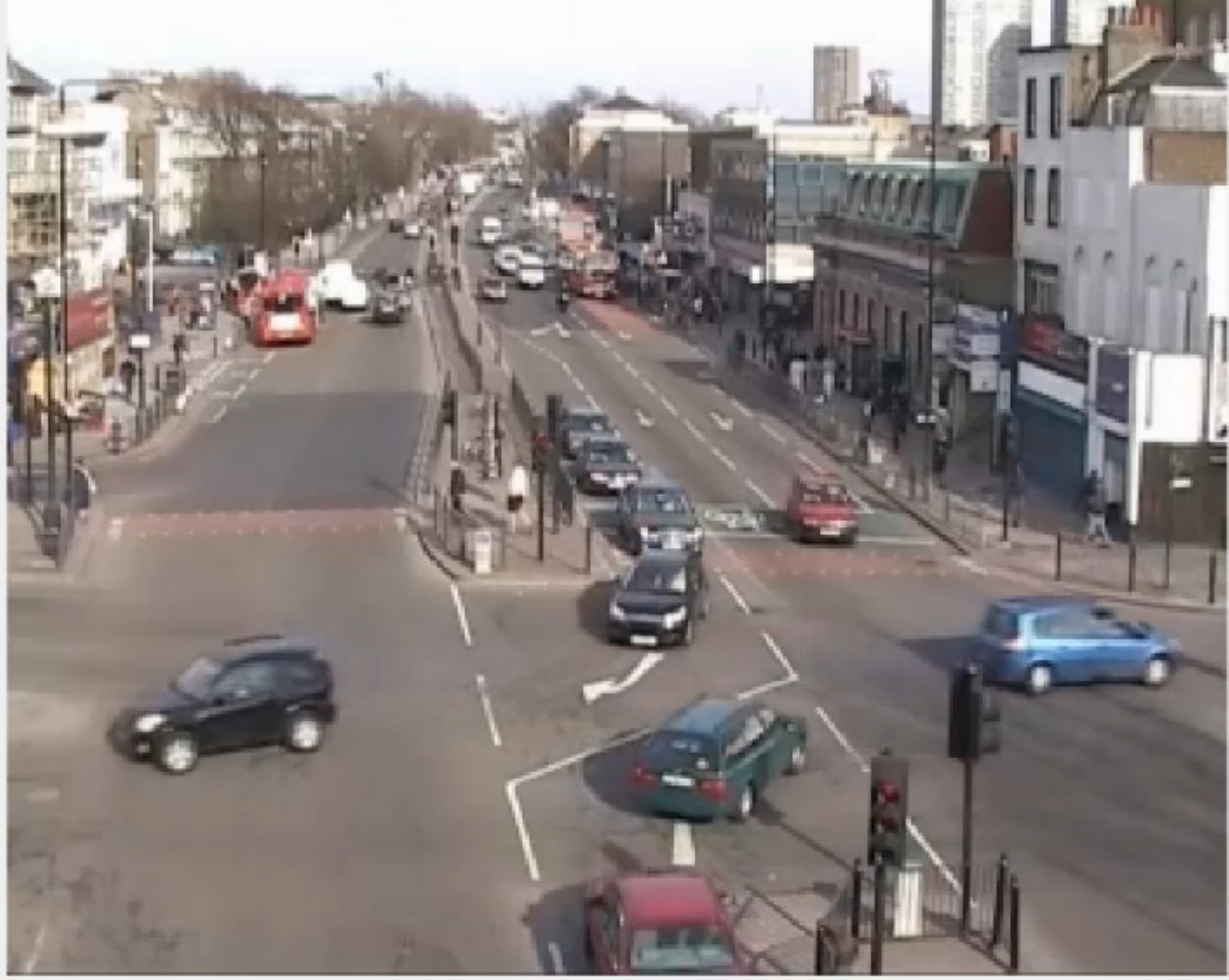}
\label{fig:qmul_normal}}%
\hfil
\subfigure[]{\includegraphics[width = 0.47\columnwidth]{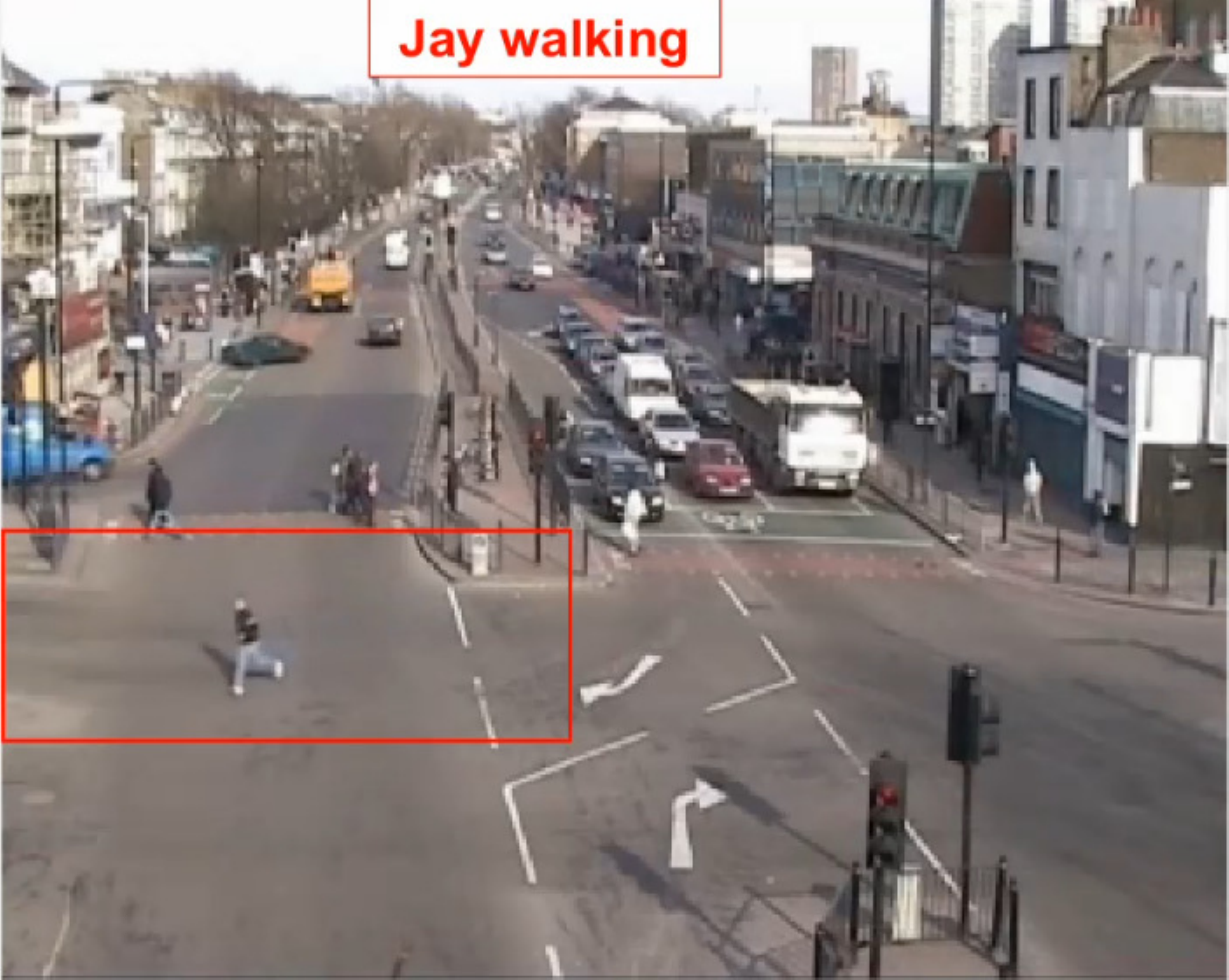}
\label{fig:qmul_jaywalking}}%
\hfil
\subfigure[]{\includegraphics[width = 0.47\columnwidth]{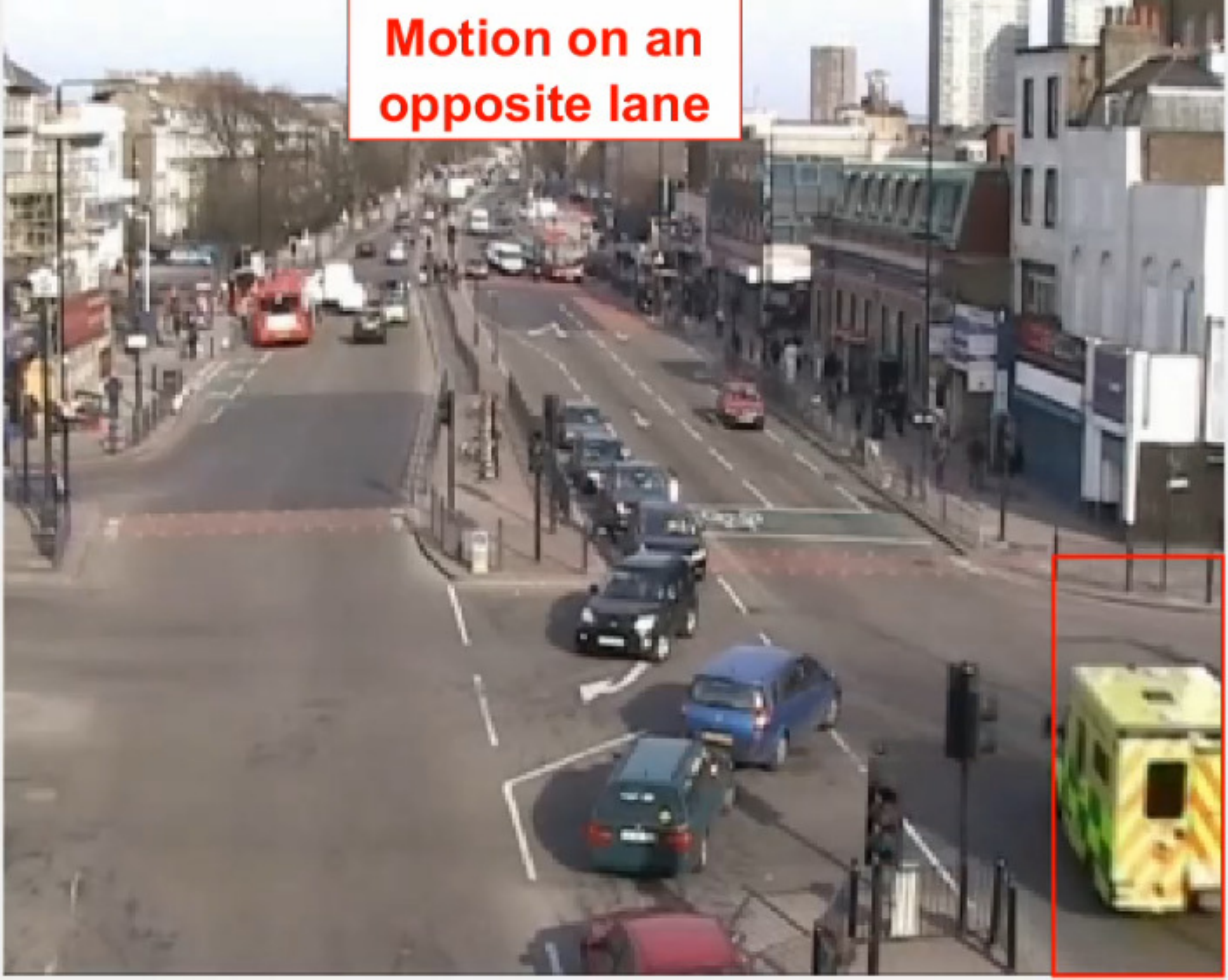}
\label{fig:qmul_wrong_direction}}%
\hfil
\subfigure[]{\includegraphics[width = 0.47\columnwidth]{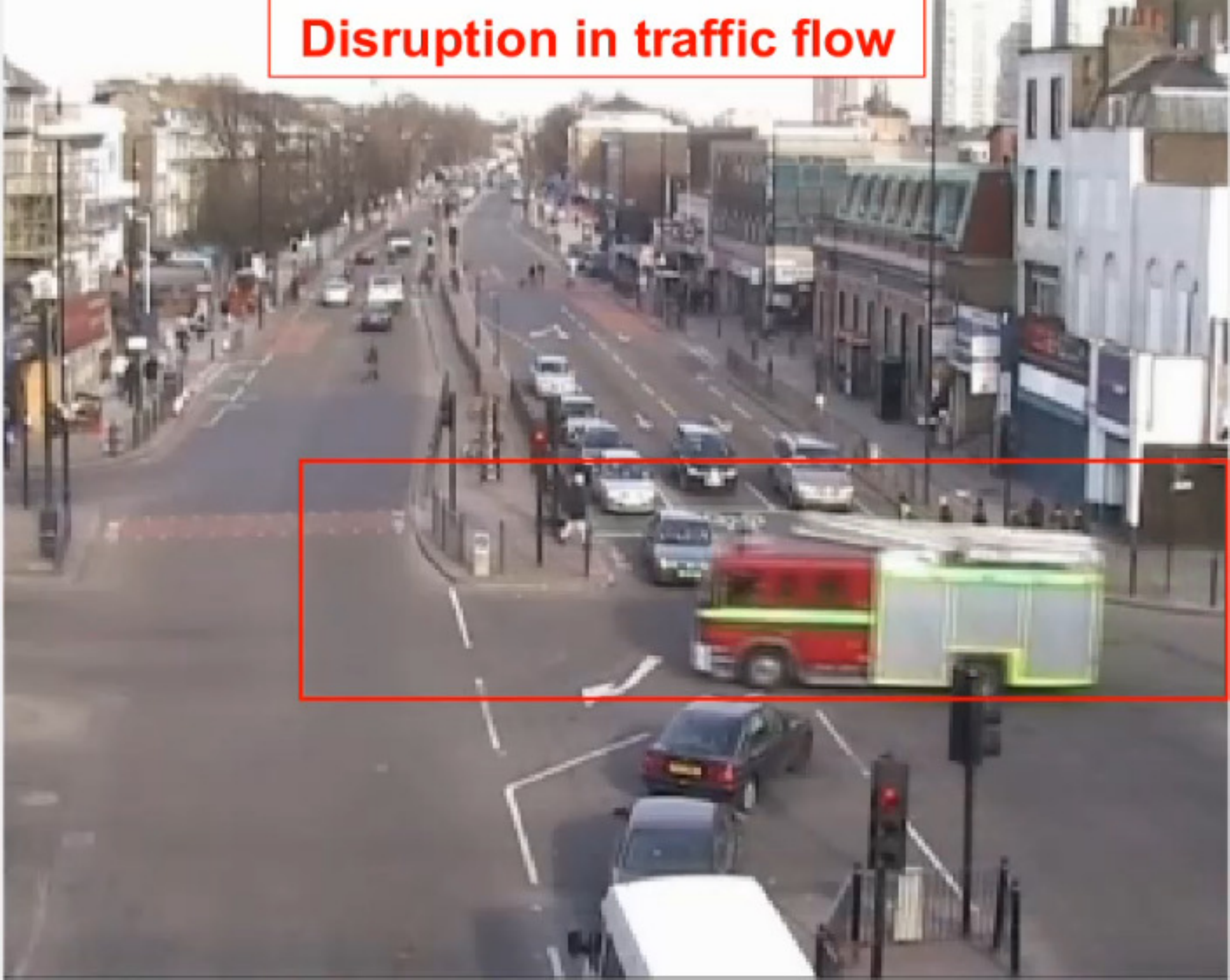}
\label{fig:qmul_disruption}}%
\caption{QMUL-junction dataset snapshots. \protect\subref{fig:qmul_normal} is an example of a normal motion, \protect\subref{fig:qmul_jaywalking} is an example of jay-walking abnormality, \protect\subref{fig:qmul_wrong_direction} is an example of a car moving on the wrong lane in the opposite to normal direction, \protect\subref{fig:qmul_disruption} is an example an emergency service car disrupting a normal traffic flow.}
\label{fig:qmul_samples}
\end{center}
\vskip -0.2in
\end{figure*}

Within the testing dataset we generate some ``abnormal'' documents where topics are chosen uniformly from the set of all the topics except those used in the previous documents. The data generated in such a way contradicts the main model assumption that the topic mixtures of the successive documents should be similar. 

Figure~\ref{fig:synthetic_roc} presents the obtained ROC-curves for anomaly detection. For the reference we also show the ROC-curve for the ``true'' model, i.e. the model with the true topics $\boldsymbol\phi_k$ and the true table and topic assignments $\mathbf{t}$ and $\mathbf{k}$. This model represents the one that can perfectly restore all the latent variables. The corresponding AUC values are in Table~\ref{tab:auc}. The proposed dynamic HDP shows the anomaly detection performance competitive to the ``true'' model.  

\subsection{Real Data}
We also test the algorithms on the QMUL-junction real video data~\cite{Hospedales2011} captured a busy road junction~(Figure~\ref{fig:qmul_normal}). 5 out of 45 minutes of the video sequence is used as a training dataset for offline Gibbs sampler. 

For the ground truth reference the data is labelled as normal and abnormal, where abnormal event examples are jay-walking (Figure~\ref{fig:qmul_jaywalking}), driving wrong direction~(Figure~\ref{fig:qmul_wrong_direction}), disruption in traffic flow~(Figure~\ref{fig:qmul_disruption}).

\begin{figure}[!t]
\vskip 0.2in
\begin{center}
\centerline{\includegraphics[scale=0.4]{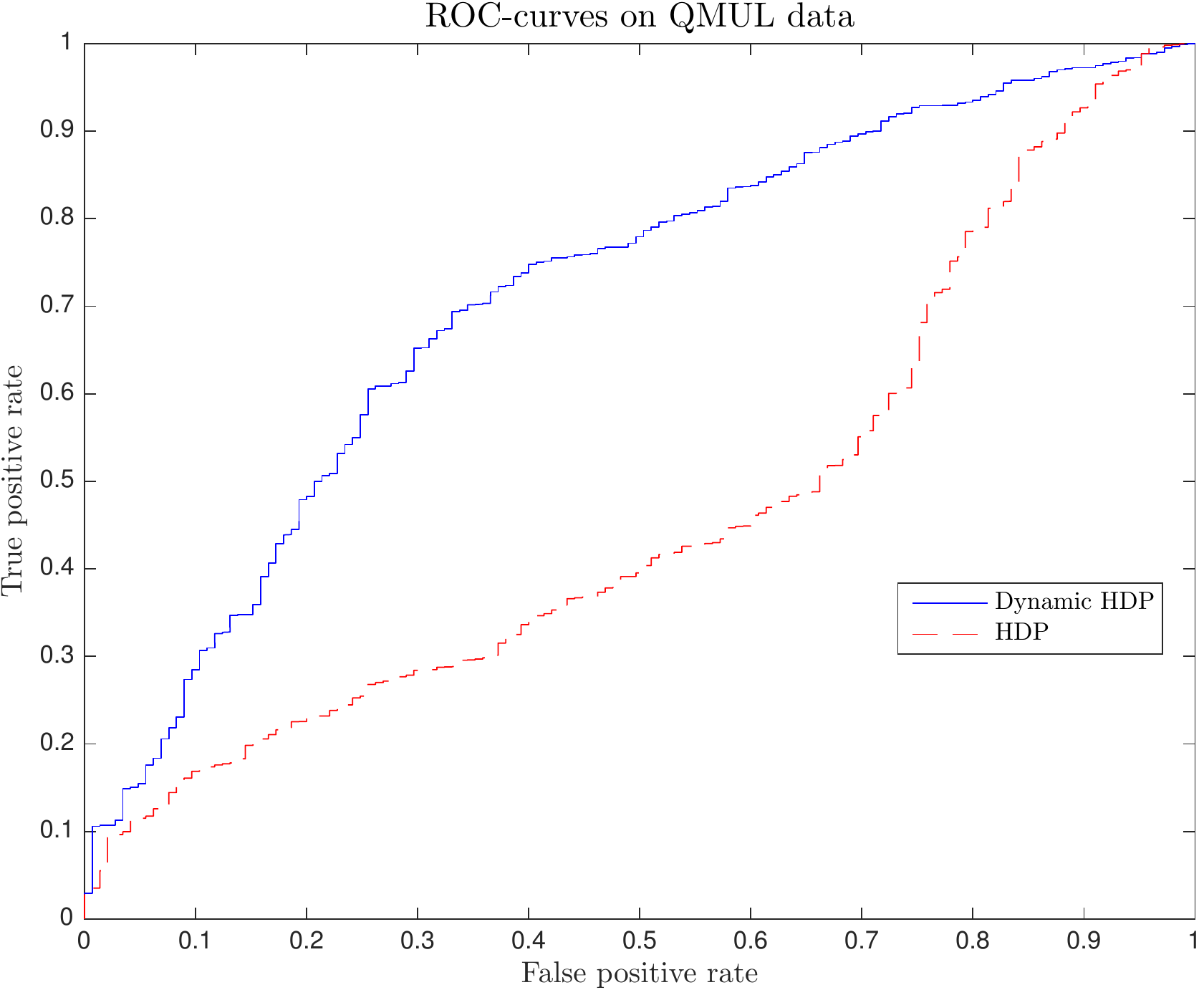}}
\caption{The ROC-curves for the QMUL data.}
\label{fig:qmul_roc}
\end{center}
\vskip -0.2in
\end{figure}

Figure~\ref{fig:qmul_roc} presents the ROC-curves while Table~\ref{tab:auc} contains the corresponding AUC values. The experiment confirms that the dynamics consideration in a topic model improves the anomaly detection performance.

\section{Conclusions}
\label{sec:conclusions}
A novel Bayesian nonparametric dynamic topic model is proposed in this paper. Batch and online inference algorithms are designed. Anomaly detection in video is considered as an application of the model for which we propose an abnormality measure. The experimental results both on synthetic and real data show that the proposed dynamic topic model improves the anomaly detection performance in comparison to the non-dynamic model.

\section*{Acknowledgments}

This work is accepted to the 19th International Conference on Information Fusion 2016. Olga Isupova and Lyudmila Mihaylova acknowledge the support from the EC Seventh Framework Programme [FP7 2013-2017] TRAcking in compleX sensor systems (TRAX) Grant agreement no.: 607400. Lyudmila Mihaylova acknowledges also the support from the UK Engineering and Physical Sciences Research Council (EPSRC) via the Bayesian Tracking and Reasoning over Time (BTaRoT) grant EP/K021516/1.

\bibliography{Biblist}

\begin{thebibliography}{17}
\providecommand{\natexlab}[1]{#1}
\providecommand{\url}[1]{\texttt{#1}}
\expandafter\ifx\csname urlstyle\endcsname\relax
  \providecommand{\doi}[1]{doi: #1}\else
  \providecommand{\doi}{doi: \begingroup \urlstyle{rm}\Url}\fi

\bibitem[Ahmed \& Xing(2010)Ahmed and Xing]{Ahmed2010}
Ahmed, Amr and Xing, Eric.
\newblock Timeline: A dynamic hierarchical {D}irichlet process model for
  recovering birth/death and evolution of topics in text stream.
\newblock In \emph{Proceedings of the Twenty-Sixth Conference Annual Conference
  on Uncertainty in Artificial Intelligence (UAI-10)}, pp.\  20--29, Corvallis,
  Oregon, 2010. AUAI Press.

\bibitem[Blei \& Lafferty(2006)Blei and Lafferty]{Blei2006Dynamic}
Blei, David~M. and Lafferty, John~D.
\newblock Dynamic topic models.
\newblock In \emph{Proceedings of the 23rd International Conference on Machine
  Learning}, ICML '06, pp.\  113--120, New York, NY, USA, 2006. ACM.

\bibitem[Blei et~al.(2003)Blei, Ng, and Jordan]{Blei03LDA}
Blei, David~M., Ng, Andrew~Y., and Jordan, Michael~I.
\newblock Latent {D}irichlet allocation.
\newblock \emph{Journal of Machine Learning Research}, 3:\penalty0 993--1022,
  March 2003.

\bibitem[Geman \& Geman(1984)Geman and Geman]{Geman1984}
Geman, Stuart and Geman, Donald.
\newblock Stochastic relaxation, {G}ibbs distributions, and the {B}ayesian
  restoration of images.
\newblock \emph{IEEE Transactions on Pattern Analysis and Machine
  Intelligence}, \penalty0 (6):\penalty0 721--741, 1984.

\bibitem[Griffiths \& Steyvers(2004)Griffiths and Steyvers]{Griffiths2004}
Griffiths, Thomas~L. and Steyvers, Mark.
\newblock Finding scientific topics.
\newblock \emph{Proceedings of the National Academy of Sciences}, 101\penalty0
  (1):\penalty0 5228--5235, 2004.

\bibitem[Hofmann(1999)]{Hofmann99}
Hofmann, Thomas.
\newblock Probabilistic latent semantic indexing.
\newblock In \emph{Proceedings of the 22nd Annual International ACM SIGIR
  Conference on Research and Development in Information Retrieval}, SIGIR '99,
  pp.\  50--57, New York, NY, USA, 1999. ACM.

\bibitem[Horn \& Schunck(1981)Horn and Schunck]{Horn1981}
Horn, Berthold~K and Schunck, Brian~G.
\newblock Determining optical flow.
\newblock \emph{Artificial Intelligence}, 17:\penalty0 185--203, 1981.

\bibitem[Hospedales et~al.(2012)Hospedales, Gong, and Xiang]{Hospedales2011}
Hospedales, Timothy, Gong, Shaogang, and Xiang, Tao.
\newblock Video behaviour mining using a dynamic topic model.
\newblock \emph{International Journal of Computer Vision}, 98\penalty0
  (3):\penalty0 303--323, 2012.

\bibitem[Jeong et~al.(2014)Jeong, Yoo, Yi, and Choi]{Jeong14}
Jeong, Hawook, Yoo, Youngjoon, Yi, Kwang~Moo, and Choi, Jin~Young.
\newblock Two-stage online inference model for traffic pattern analysis and
  anomaly detection.
\newblock \emph{Machine Vision and Applications}, 25\penalty0 (6):\penalty0
  1501--1517, 2014.

\bibitem[Kuettel et~al.(2010)Kuettel, Breitenstein, Van~Gool, and
  Ferrari]{Kuettel2010}
Kuettel, D., Breitenstein, M.D., Van~Gool, L., and Ferrari, V.
\newblock What's going on? {D}iscovering spatio-temporal dependencies in
  dynamic scenes.
\newblock In \emph{Proceedings of the 2010 IEEE Conference on Computer Vision
  and Pattern Recognition (CVPR)}, pp.\  1951--1958, June 2010.

\bibitem[Mehran et~al.(2009)Mehran, Oyama, and Shah]{Mehran09}
Mehran, Ramin, Oyama, Alexis, and Shah, Mubarak.
\newblock {Abnormal crowd behavior detection using social force model}.
\newblock In \emph{Proceedings of the 2009 IEEE Conference on Computer Vision
  and Pattern Recognition}, pp.\  935--942. IEEE, June 2009.

\bibitem[Murphy(2012)]{Murphy2012}
Murphy, Kevin~P.
\newblock \emph{Machine learning: a probabilistic perspective}.
\newblock MIT press, 2012.

\bibitem[Pruteanu-Malinici et~al.(2010)Pruteanu-Malinici, Ren, Paisley, Wang,
  and Carin]{PruteanuMalinici2010}
Pruteanu-Malinici, I., Ren, Lu, Paisley, J., Wang, E., and Carin, L.
\newblock Hierarchical {B}ayesian modeling of topics in time-stamped documents.
\newblock \emph{IEEE Transactions on Pattern Analysis and Machine
  Intelligence}, 32\penalty0 (6):\penalty0 996--1011, June 2010.

\bibitem[Srebro \& Roweis(2005)Srebro and Roweis]{Srebro2005}
Srebro, N. and Roweis, S.
\newblock Time-varying topic models using dependent {D}irichlet processes.
\newblock Technical report, Technical Report, Departament of computer science,
  University of Toronto, 2005.

\bibitem[Teh et~al.(2006)Teh, Jordan, Beal, and Blei]{Teh2012}
Teh, Yee~Whye, Jordan, Michael~I, Beal, Matthew~J, and Blei, David~M.
\newblock Hierarchical {D}irichlet processes.
\newblock \emph{Journal of the American Statistical Association}, 101\penalty0
  (476):\penalty0 1566--1581, 2006.

\bibitem[Varadarajan \& Odobez(2009)Varadarajan and Odobez]{Varadarajan2009}
Varadarajan, J. and Odobez, J.
\newblock Topic models for scene analysis and abnormality detection.
\newblock In \emph{Proceedings of the 2009 IEEE 12th International Conference
  on Computer Vision Workshops (ICCV Workshops)}, pp.\  1338--1345, Sept 2009.

\bibitem[Zhang et~al.(2010)Zhang, Song, Zhang, and Liu]{Zhang2010}
Zhang, Jianwen, Song, Yangqiu, Zhang, Changshui, and Liu, Shixia.
\newblock Evolutionary hierarchical {D}irichlet processes for multiple
  correlated time-varying corpora.
\newblock In \emph{Proceedings of the 16th ACM SIGKDD international conference
  on Knowledge discovery and data mining}, pp.\  1079--1088. ACM, 2010.

\end{thebibliography}
\bibliographystyle{icml2016}

\end{document}